\documentclass{article}

\usepackage{arxiv}

\usepackage[utf8]{inputenc} 
\usepackage[T1]{fontenc}    

\usepackage[colorlinks=true, allcolors=blue]{hyperref}
\newcommand{\email}[1]{\href{mailto:#1}{\tt{\nolinkurl{#1}}}}
\newcommand{\orcid}[1]{ORCID: \href{https://orcid.org/#1}{\tt{\nolinkurl{#1}}}}

\usepackage{url}            
\usepackage{booktabs}       
\usepackage{amsfonts}       
\usepackage{nicefrac}       
\usepackage{microtype}      
\usepackage{lipsum}
\usepackage{amssymb}
\usepackage{latexsym}

\usepackage{url}
\usepackage{xcolor}
\usepackage{hyperref}

\usepackage{amsmath}
\usepackage{graphicx}
\usepackage{cleveref}
\usepackage[export]{adjustbox}
\usepackage{subfig}
\usepackage{mwe}
\usepackage{graphicx}
\usepackage{pdflscape}

\title{Detecting and Counting Pistachios based on Deep Learning}

\author{
 Mohammad Rahimzadeh \\
  School of Computer Engineering\\
  Iran University of Science and Technology, Iran\\
  \texttt{\email{mr7495@yahoo.com}} \\
  \texttt{\orcid{0000-0002-8550-8967}} \\
  \texttt{Corresponding author}
  \And
 Abolfazl Attar \\
  Department of Electrical Engineering\\
  Sharif University of Technology, Iran\\
  \texttt{\email{attar.abolfazl@ee.sharif.edu}} \\
  \texttt{\orcid{0000-0001-6727-432X}}}

\begin{document}
\maketitle

\begin{abstract}
Pistachios are nutritious nuts that are sorted based on the
shape of their shell into two categories: Open-mouth
and Closed-mouth. The open-mouth pistachios are
higher in price, value, and demand than the closed-mouth pistachios. Because of these differences, it is considerable for production companies to precisely count the number of each kind. This paper aims to propose a new system for counting the different types of pistachios with computer vision. 
We have introduced and shared a new dataset of pistachios, including six videos with a total length of 167 seconds and 3927 labeled pistachios.
Unlike many other works, our model counts pistachios in videos, not images. Counting objects in videos need assigning each object between the video frames so that each object be counted once. The main two challenges in our work are the existence of pistachios' occlusion and deformation of pistachios in different frames because open-mouth pistachios that move and roll on the transportation line may appear as closed-mouth in some frames and open-mouth in other frames. 
Our novel model first is trained on the RetinaNet object detector network using our dataset to detect different types of pistachios in video frames. After gathering the detections, we apply them to a new counter algorithm based on a new tracker to assign pistachios in consecutive frames with high accuracy. Our model is able to assign pistachios that turn and change their appearance (e.g., open-mouth pistachios that look closed-mouth) to each other so does not count them incorrectly.
 Our algorithm performs very fast and achieves good counting results. The computed accuracy of our algorithm on six videos (9486 frames) is 94.75\%.
\end{abstract}


\section{Introduction}
\label{1}
Nowadays, automation in the industry plays a significant role in increasing efficiency and saving resources. One of the industries that need more development in automation than other industries is the agricultural industry and related fields. Proper packaging of agricultural products will increase profitability and reduce crop losses. On the other hand, crop quality categorization depends on human resources, which causes time-consuming and rising costs, and most importantly, does not have the necessary quality compared to machines.

Pistachio is one of the crops that need human resources to classify and count so that the quality of the crop can be evaluated in terms of its open or closed shell. Pistachios are mostly sorted based on their shell's shape to open-mouth and closed-mouth, and these two kinds differ in price and value.

Pistachios are used as nuts, and in the food industry \cite{woodruff1979tree}. Pistachio kernels are rich in unsaturated fatty acids, fiber, carbohydrates, proteins, and various vitamins that are very useful for the human diet \cite{maskan1998fatty, kashaninejad2006some}. Adequate consumption of pistachio kernels reduces the risk of heart disease and has a good effect on blood pressure in people who do not have diabetes, and prevents some cancers\cite{food2003qualified, kashaninejad2006some, wikipedia_2020}.
Pistachio is one of Middle Eastern countries' main agricultural products, especially Iran, \cite{faostat}. The largest producers of this product in the world are Iran, the USA, and Turkey, respectively \cite{faostat}.

\begin{figure*}[!ht]
\centering
\includegraphics[scale=0.6]{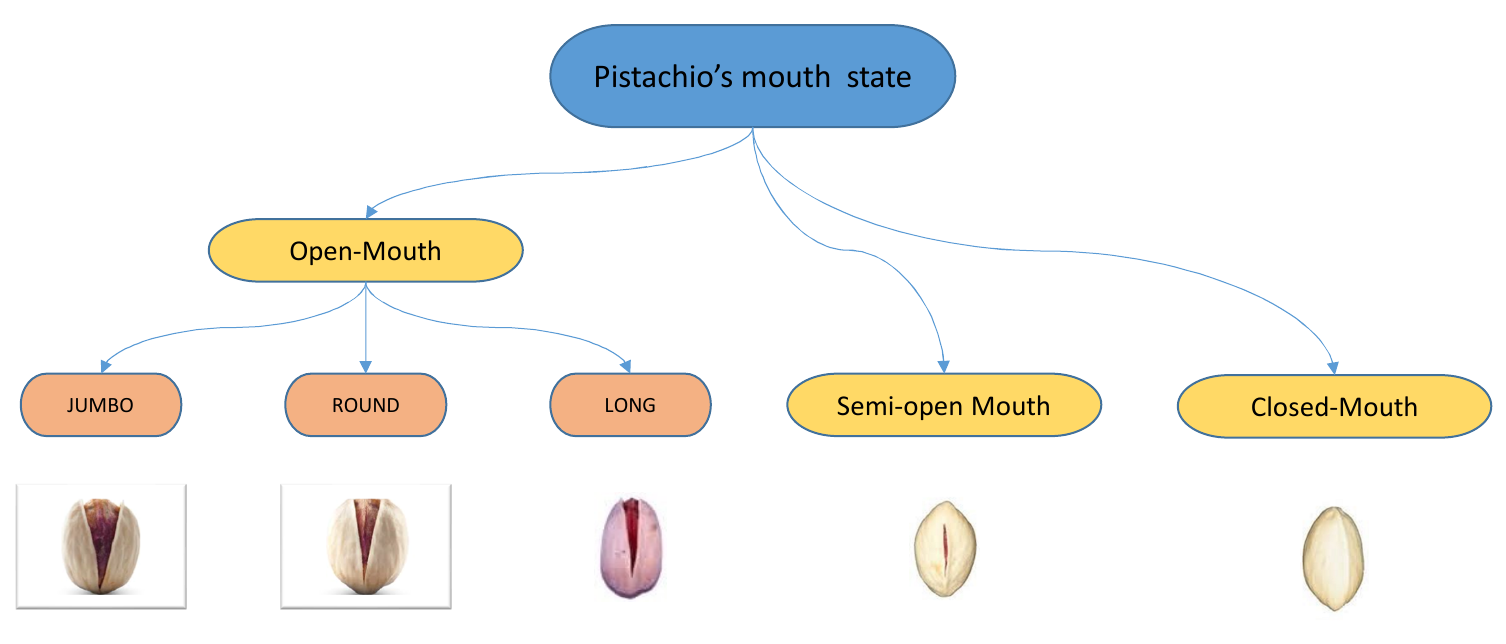}
\caption{Pistachios Assortment}
\label{fig1}
\end{figure*}

There are many types of pistachios, depending on the type and place of growth; they have different sizes, colors, and flavors \cite{ratinkhosh_2020}. Depending on the shape of the pistachio, it can be divided into three general categories: round, long, and jumbo \cite{ratinkhosh_2020}. Long pistachios have a narrower split than the other two, and the round and jumbo type have a much clearer split than the semi-closed one  \cite{ratinkhosh_2020}.
Fig. \ref{fig1} shows a summary of pistachios' different types.

The average weight of each pistachio's shell is about 0.57 grams \cite{kiger_2017}, which is about 1750 per kilogram. As a result, according to the statistics provided, counting them is a very time-consuming and tedious task that artificial intelligence can easily do.

Detecting and counting the pistachios can be used for proper packaging and crop quality assessment. Another advantage of this can be the estimation of the amount of the crops in the coming years and the breeding of pistachio trees to increase the quality of the crop. As there is a significant difference in price and demand between the open-mouth and closed-mouth pistachios, the factories related to pistachio production or packaging need to know precisely how much of these two kinds exist in every package. Counting these two kinds of pistachios can also help separate them to increase the exporting packages' quality. Counting the pistachios by human resources is very time-consuming and practically uneconomical, so machine vision can play a significant role in this regard.
 
Another application of these procedures is that closed pistachios are used by the mechanical opening method \cite{burlock1991apparatus} to be returned to the consumption cycle. In this method, it is first necessary to identify the closed pistachios, which results in reduced losses and increased crop yields \cite{burlock1991apparatus}.
 
One of the new methods for detecting, counting, and classifying pistachios is machine vision. In recent years, machine vision has been used for many tasks to automate and replace machines with humans, which have yielded excellent results \cite{shrestha2019review}. These applications exist in the fields of medicine \cite{lee2017deep}, medical image diagnosis \cite{rahimzadeh2020modified,rahimzadeh2020fully}, self-driving \cite{garcia2017review}, security \cite{berman2019survey}, and agriculture \cite{rahnemoonfar2017deep, liu2018robust, murecsan2018fruit}, and so on. 

One of the principles of using robots and remote control and sense is the use of machine vision. Therefore, improving the accuracy and precision of the system is one of the essential principles. In machine vision, various methods such as thermal cameras, sensors, microscopes, and common cameras have been used for imaging space around it. However, the main issue in machine vision is the choice of the data analysis method. 

Currently, one of the most attractive and accurate methods of machine vision is deep learning, which has been created a revolution in artificial intelligence \cite{lecun2015deep}.
One of the most important advantages of deep convolutional neural networks is the comprehensive and flexible recognition of different objects. \cite{krizhevsky2012imagenet}. 

Using the deep convolutional neural networks, we can identify and count pistachios. The appearance of the pistachio, the angle of the camera or robot also plays a vital role in correctly determining the open and closed pistachios.

Our work is the first to investigate the detecting, tracking, and counting performance of pistachios on transportation lines, even in high density and occlusion situations.  Another critical challenge is to count the open-mouth and closed-mouth pistachios correctly because the open-mouth pistachios can show themselves as closed-mouth pistachio when moving and spinning on the transportation line. Based on this situation, many tracking algorithms will fail because the class type of pistachio changes between successive frames. Still, our tracking and counting model has solved this problem and achieves high accuracy.

\begin{figure*}[!hb]
\centering
\includegraphics[scale=0.67]{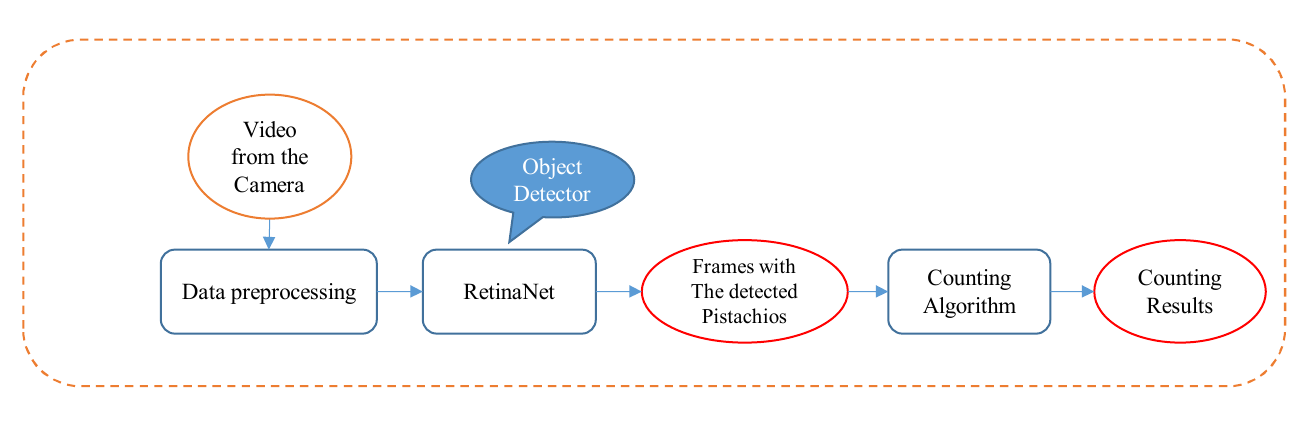}
\caption{General Schematic of our proposed model}
\label{general}
\end{figure*}

In this paper, at first, we will propose our dataset, which we call Pesteh-Set. At the next stage, we will describe the detection phase. We have implemented the RetinaNet \cite{lin2018focal}  object detector for detecting the Pistachios in video frames. We have separated the dataset into five-folds and allocated 20 percent of the dataset for testing and the rest for the training. After the detection phase, we present the model for counting the open-mouth and closed-mouth pistachios. This algorithm runs very fast with high accuracy. The general schematic of our work is presented in Fig. \ref{general}.

\begin{figure*}[!hb]
\centering
\includegraphics[scale=0.5]{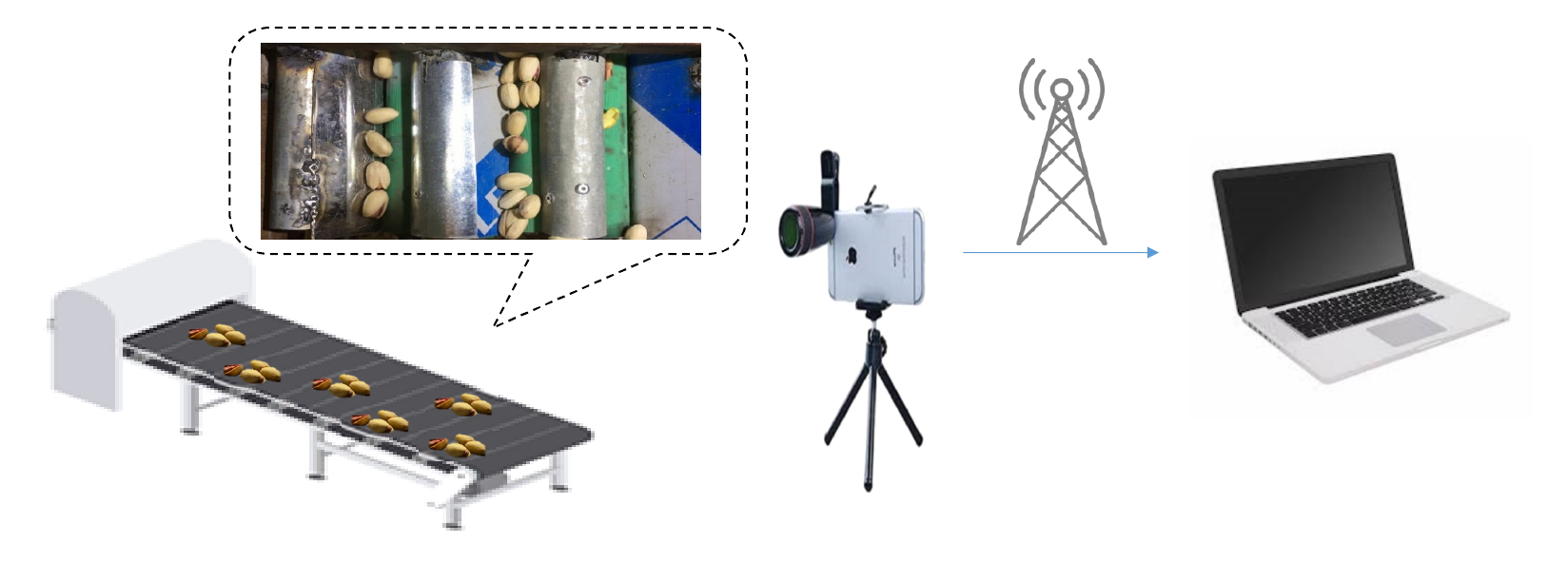}
\caption{The General View of how Pesteh-Set was recoreded and our proposed system for counting the pistachios}
\label{fig3}
\end{figure*}

One of the closest studies in our research is fruit detection and counting \cite{gongal2015sensors}. This is done using a variety of tools such as a B/W camera \cite{whittaker1987fruit}, a color camera \cite{cohen2010estimation}, a thermal camera \cite{stajnko2004estimation} and a spectral camera \cite{safren2007detection}. Due to the type of data we have, color is one of the most important features, B/W camera is not suitable \cite{whittaker1987fruit}. On the other hand, because the spectral camera has a time delay, this method cannot be suitable \cite{safren2007detection}. Due to its sensitivity to size and lack of detection of split pistachios, the thermal camera is not suitable for analyzing our data \cite{stajnko2004estimation}. As a result, the color camera is more suitable than other tools. Another advantage of using color cameras is their abundance, especially in mobile phones, which can be used for remote monitoring and control. 

There are other methods, such as sensors, that can be used for counting, but since one of the challenges ahead is split and non-split pistachios counting, however, this method is not suitable for our data \cite{feng2012novel}.

One of the most widely used classifications and detection methods used is the K-means clustering method, which is performed unsurprisingly. In \cite{wachs2010low}, K-means clustering detects green apples using thermal and color cameras. 

One of the most basic methods of Supervised classification is the Bayesian classifier, which has been used in \cite{slaughter1989discriminating} to identify oranges and has yielded relatively good results from previous research. Other used methods include KNN clustering \cite{linker2012determination}.

Artificial neural networks have a special place in machine vision and object detection. In the meantime, deep convolutional neural networks have shown excellent results for images and videos, too. Therefore, image detection systems move to increase the accuracy and quality of deep learning. 

In \cite{murecsan2018fruit}, they classify different fruits using an innovative deep neural network. In \cite{rahnemoonfar2017deep}, a deep neural network was developed to detect and count the number of tomatoes per plant. In this study, due to the lack of enough data, the data were generated by simulating a green and brown environment in which the tomatoes were simulated with red circles. This can take the results away from the real world. 

In \cite{liu2018robust}, the environment is photographed using a monocular camera, and visible fruits are detected and tracked on the tree. This detection is made by training a fully Convolutional Network, then using image processing methods to track the fruits and then counting them.

\cite{rahimzadeh2020sperm} is one of the papers that has done very well in detecting and tracking motile objects. This paper also introduced a method for improving motile-objects detection.
In this article, by using RetinaNet \cite{lin2018focal}, and other introduced methods, they have detected motile sperms in the video frames. Finally, by implementing a new tracker called modified CSR-DCF, they have tracked and analyzed the sperms attributes, such as their number and motility characteristics.

The rest of the paper is organized as follows: In section \ref{3}, we describe our dataset and the detection and counting models. In section \ref{4}, the results of our work are presented. In section \ref{5}, we discuss the obtained results, and in section \ref{6}, the paper is concluded.

\section{Materials and methods}
\label{3}
\subsection{Pesteh-Set}
\label{31}
Pistachio is known as Pesteh in Iran; that is why we called our dataset Pesteh-Set. 
Pesteh-Set \footnote{<This dataset is shared in \url{https://github.com/mr7495/Pesteh-Set}>} is made of two parts. The first part includes 423 images with ground truth. We sorted the pistachios into two classes: Open-mouth and closed-mouth. The images' ground truth consists of the bounding boxes of the two classes of pistachios in the images.
There are between 1 to 27 pistachios in each image and 3927 pistachios totally. The second part includes six videos (9486 frames) that were used for the counting phase. These six videos include 561 motile pistachios and more than 350,000 single pistachios (sum of pistachios in all frames).

The videos of the dataset have been recorded by a cell-phone camera with 1920 × 1080 pixels resolution. Five of these videos are recorded with 60 frames per second(fps) frame rate, and one other is recorded with 30 fps frame rate. The cell-phone was perched on the wall above the line that was transporting the pistachios. This line was designed somehow that the pistachios could roll on it. The reason the pistachios rolling is so important is that the open-mouth pistachios could appear on their backside where they look like closed-mouth pistachios, but the rolling cause them to show their open-mouth side when rolling. Fig. \ref{fig3} presents a view of how the dataset was recorded and the general schematic of our proposed system for remote counting the pistachios.

\begin{figure}[!ht]
\centering
\includegraphics[width=0.49\linewidth]{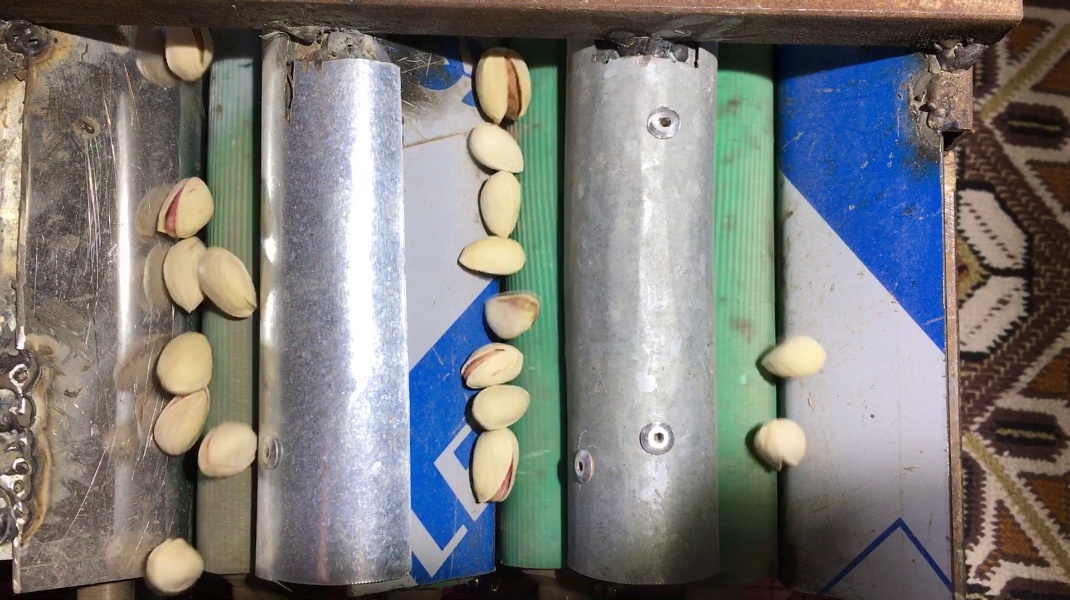}
\includegraphics[width=0.49\linewidth]{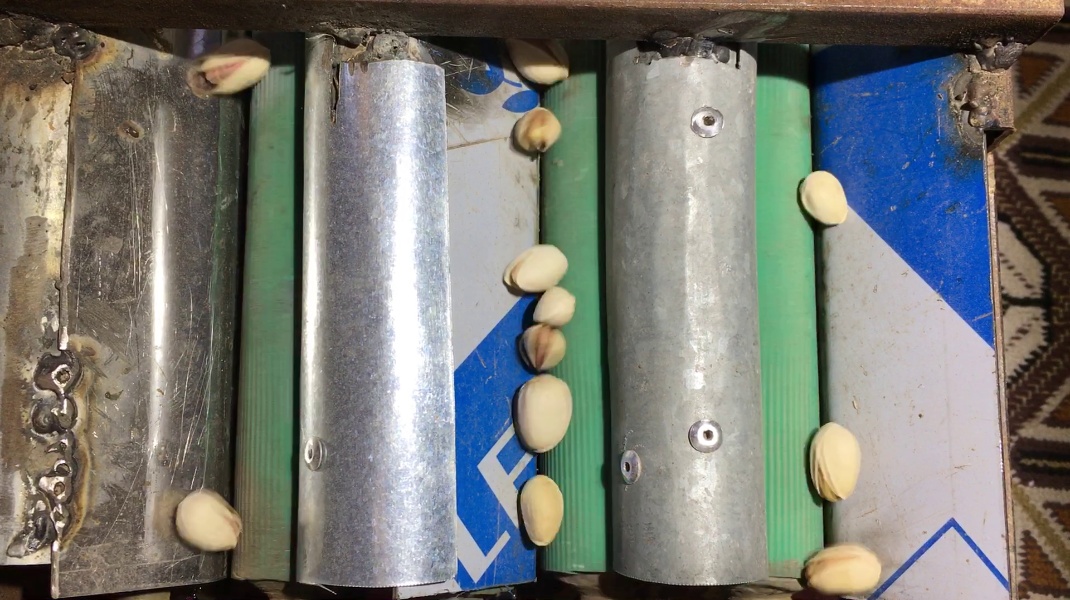}

\includegraphics[width=0.49\linewidth]{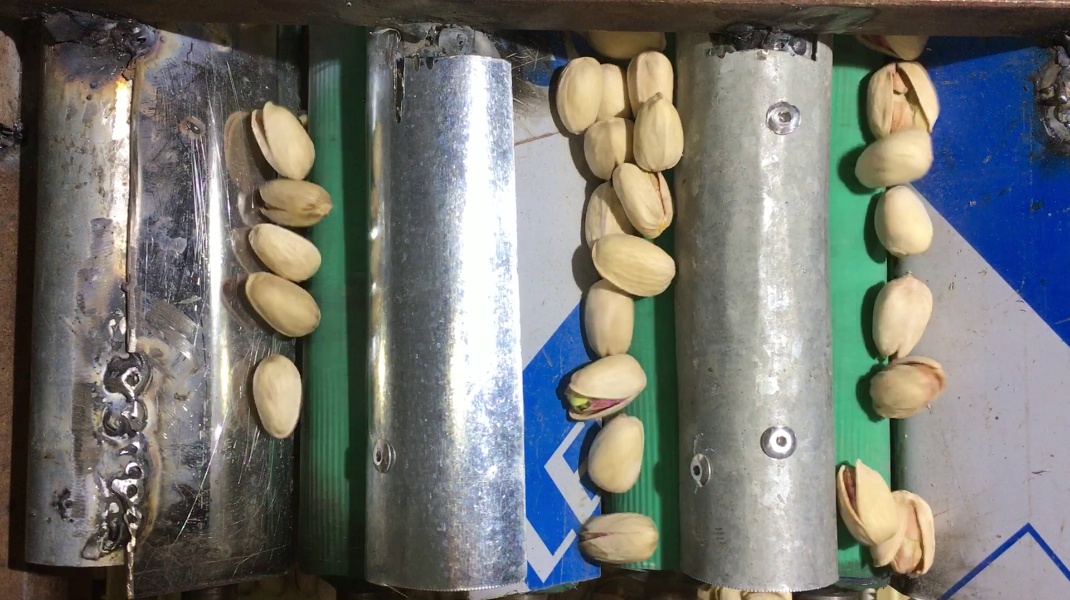}
\includegraphics[width=0.49\linewidth]{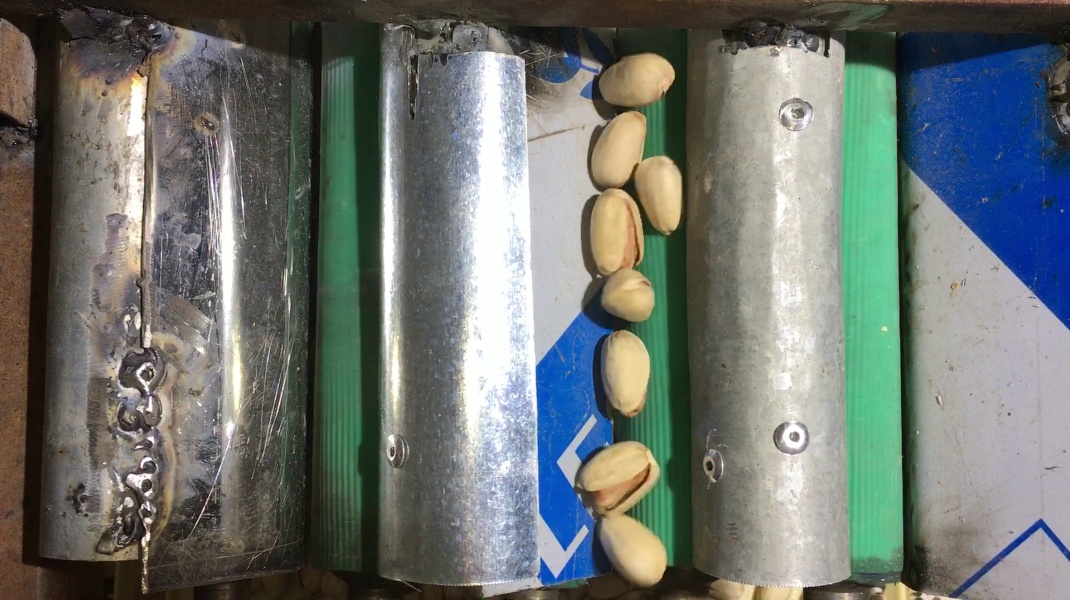}
\caption{Some of the images in Pesteh-Set}
\label{fig9}
\end{figure}

We have selected some frames of the videos and labeled them with a self-developed program using OpenCV library \cite{2015opencv} on python language. 
The pistachios are categorized into two classes: open-mouth pistachios and closed-mouth pistachios. Some of the images of this dataset are presented in Fig. \ref{fig9}.
The self-developed program for labeling the images along with all the data has been shared so other researchers could use them to make the Pistachio-Dataset larger. Table. \ref{table1} presents the details of Pesteh-Set.

\begin{table*}[!ht]
\centering
\large
\caption{This table shows the distribution of Pistachios in Pesteh-Set}

\begin{tabular}{l|l|l|l|}
\cline{2-4}
                                                                                              & \begin{tabular}[c]{@{}l@{}}Number of \\ Open-mouth\\ Pistachios\end{tabular} & \begin{tabular}[c]{@{}l@{}}Number of \\ Closed-mouth\\ Pistachios\end{tabular} & \begin{tabular}[c]{@{}l@{}}Number\\ of All\\ the Pistachios\end{tabular} \\ \hline
\multicolumn{1}{|l|}{Video 1}                                                                 & 50                                                                           & 20                                                                             & 70                                                                       \\ \hline
\multicolumn{1}{|l|}{Video 2}                                                                 & 60                                                                           & 20                                                                             & 80                                                                       \\ \hline
\multicolumn{1}{|l|}{Video 3}                                                                 & 70                                                                           & 20                                                                             & 90                                                                      \\ \hline
\multicolumn{1}{|l|}{Video 4}                                                                 & 90                                                                           & 20                                                                             & 110                                                                      \\ \hline
\multicolumn{1}{|l|}{Video 5}                                                                 & 100                                                                          & 20                                                                             & 120                                                                      \\ \hline
\multicolumn{1}{|l|}{Video 6}                                                                 & 39                                                                           & 52                                                                             & 91                                                                       \\ \hline
\multicolumn{1}{|l|}{\begin{tabular}[c]{@{}l@{}}All of the\\ Videos\end{tabular}}             & 409                                                                          & 152                                                                            & 561                                                                      \\ \hline
\multicolumn{1}{|l|}{\begin{tabular}[c]{@{}l@{}}All the 423\\ labeled \\ images\end{tabular}} & 1993                                                                         & 1934                                                                           & 3927                                                                     \\ \hline

\end{tabular}
\label{table1}
\end{table*}

In Table. \ref{table1}, the reason that the number of pistachios in the videos is less than the images is that the number of pistachios in the videos denotes the number of mobile pistachios. It means, from the moment one pistachio enters the video in a frame until it exits the video in the later frames, it would be counted as one pistachio.

To report the number of pistachios in the images, we counted the number of pistachios in each image. It is noteworthy that we selected non-consecutive frames of different videos for labeling, so there would not be a similarity between them (for training the models). Besides, we tried to choose the frames somehow that we have an almost equal number of each class, and our dataset become balanced for training.

\begin{figure*}[!ht]
\centering
\includegraphics[width=\linewidth]{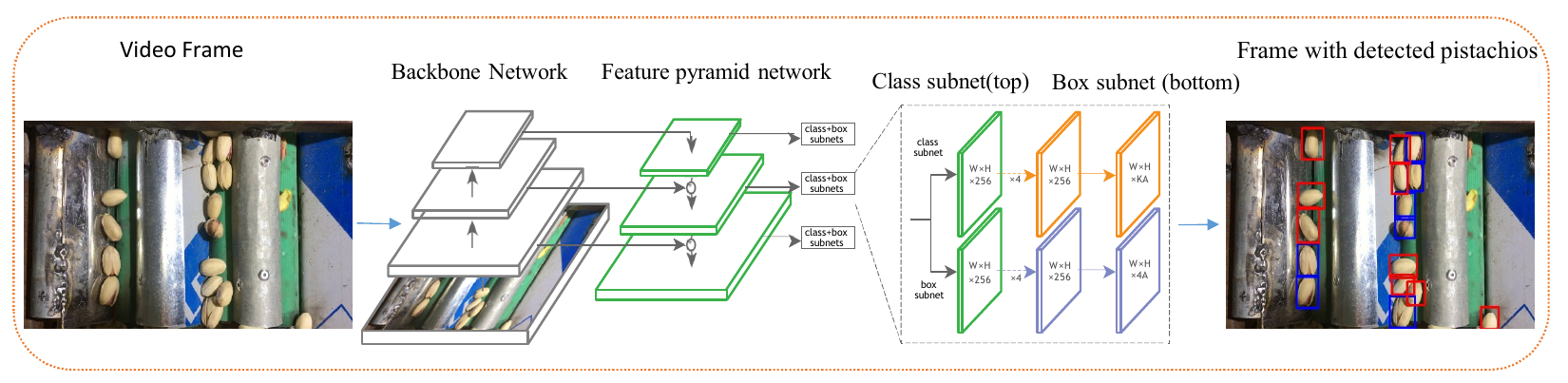}
\caption{RetinaNet Architecture}
\label{fig5}
\end{figure*}

\subsection{Detection}
\label{32}
\subsubsection{RetinaNet}
\label{321}
RetinaNet \cite{lin2018focal} is a deep fully convolutional neural network that is utilized for object detection. The architecture of RetinaNet is depicted in Fig. \ref{fig5}.
RetinanNet is made of three main parts. The first part, which is the feature extractor, is build up from a backbone model and the feature pyramid network (FPN) \cite{lin2017feature}. Famous convolutional networks like ResNet \cite{he2016deep}, DenseNet \cite{huang2017densely}, and VGG \cite{simonyan2014very} are mostly used as the backbone of RetinaNet. The FPN takes the multi-dimensional features that are extracted from the backbone network as the input to build a multi-scale feature pyramid from the input image \cite{lin2017feature}. The usage of the FPN on top of the backbone considerably improves object detection accuracy because it makes the model able to detect multi-scale objects.

The second part of the RetinaNet is the Classifier, which has the role in predicting the possibility of the presence of each of the classes at each spatial location for each of the anchor boxes. The third part is the box regression that regresses each of the anchor boxes to the nearest ground truth object boxes \cite{lin2018focal}. Another novelty presented in the RetinaNet is using the focal loss \cite{lin2018focal} as the loss function. The focal loss adds a modulating factor to the cross-entropy loss function, and by doing so, it focuses on the hard examples while training, and as the loss of hard examples is higher than the easy examples, it improves the learning process and accuracy.

\subsubsection{Training}
\label{322}
We have separated the Pesteh-Set into five folds for training, which in each fold, 20 percent of the dataset was allocated for testing, and the rest for training.  The original size of images in our dataset is 1070x600 pixels, but they were preprocessed and then resized to 1333×747 pixels for training and testing the object detectors.

We used RetinaNet \cite{lin2018focal}, as the object detector. We trained and validated RetinaNet on 3 different backbones: ResNet50 \cite{he2016deep}, ResNet152 \cite{he2016deep}, and VGG16 \cite{simonyan2014very}. VGG16 \cite{simonyan2014very} was introduced by Oxford researchers and in 2014 and is made of 13 layers for the feature extraction part and represents a feature map with 512 channels. ResNet \cite{he2016deep} series was introduced after the VGG and introduced residual layers that improved deep convolution networks. These layers help the networks avoid losing data when the features being corrupted while passing through layers by applying skip connections from the previous layers to the next layers. Since ResNet, most of the next model also used and inspired from residual layers. ResNet comes with several models like ResNet50, ResNet101, and ResNet152; the difference between them is having more layers, and all of them represent a feature map with 2048 channels. ResNet152 achieved higher results on the ImageNet dataset \cite{deng2009imagenet}. The details of RetinaNet on these backbone models are described in Table. \ref{param}.

\begin{table*}[!ht]
\centering
\large
\caption{The number of parameters and training, and inference time of RetinaNet on different backbones are listed in this table.}
\begin{tabular}{|l|l|l|l|}
\hline
Backbone  & Number of Parameters & Training Time per Epoch & \begin{tabular}[c]{@{}l@{}}Inference Time per Image\\ with Size (1333,747,3)\end{tabular} \\ \hline
VGG16     & 23,428,470           & 318 s                   & 17.44 ms                                                                                  \\ \hline
ResNet50  & 36,403,702           & 390 s                   & 17.80 ms                                                                                  \\ \hline
ResNet152 & 71,137,782           & 615 s                   & 20.01 ms                                                                                  \\ \hline
\end{tabular}
\label{param}
\end{table*}

Transfer learning from the ImageNet \cite{deng2009imagenet} pre-trained weights was utilized on the backbone model at the beginning of the training to speed up the network convergence. We also used data augmentation methods like rotation, translation, shearing, horizontal and vertical flipping, and rescaling to improve the learning efficiency and stop the network from overfitting.

We applied Adam optimizer with a 1e-5 initial learning rate and an automatic reduction function to reduce the learning rate when training loss cannot be improved for five epochs. The focal loss was used for the classification subnet, and the Smooth L1 loss function was used for the regression subnet. 

In classification problems, it is most usual to set batch size between 32 to 256, so the model can analyze several images in each step of training and learn the features better. In object detection tasks, if several objects are available in each image, even if the batch size is equivalent to 1, the model analyzes several objects in each step of training. That means the actual batch size is equal to the number of objects in that image. Based on this fact, as there are several pistachios in each image, we set the batch size equal to 1 to reduce computational costs.

The applied training parameters are listed in the Table. \ref{table2} and the details of each fold are present in Table. \ref{table3}.

\begin{table*}[!ht]
\centering
\large
\caption{This table shows all the parameters and methods we used in training}
\begin{tabular}{|l|l|}
\hline
Training Parameters          & Value                                                                                                                  \\ \hline
Learning Rate                & 1e-5 (With automatic reduction  based on Loss value)                                                                    \\ \hline
Batch Size                   & 1                                                                                                                      \\ \hline
Optimizer                    & Adam                                                                                                                   \\ \hline
Loss Function                & \begin{tabular}[c]{@{}l@{}}Focal Loss for the classification subnet\\ Smooth L1 for the regression subnet\end{tabular} \\ \hline
Steps                        & 1017                                                                                                                   \\ \hline
Horizontal/Vertical flipping & Yes (50\%)                                                                                                             \\ \hline
Translation Range            & -0.1 - 0.1                                                                                                             \\ \hline
Rotation Range               & 0 - 360 degree                                                                                                         \\ \hline
Shear Range                  & -0.1 - 0.1                                                                                                             \\ \hline
Scaling Range                & -0.1 - 0.1                                                                                                             \\ \hline
\end{tabular}
\label{table2}
\end{table*}

\begin{table*}[!hb]
\centering
\large
\caption{This table presents the details of our train and test sets in each fold}
\begin{tabular}{|l|l|l|l|l|l|l|}
\hline
\begin{tabular}[c]{@{}l@{}}Fold\\ Number\end{tabular} & \begin{tabular}[c]{@{}l@{}}Training \\ Images\end{tabular} & \begin{tabular}[c]{@{}l@{}}Test \\ Images\end{tabular} & \begin{tabular}[c]{@{}l@{}}Open-mouth\\ Pistachios in\\ Training Set\end{tabular} & \begin{tabular}[c]{@{}l@{}}Closed-mouth\\ Pistachios in\\ Training Set\end{tabular} & \begin{tabular}[c]{@{}l@{}}Open-mouth\\ Pistachios in\\ Testing Set\end{tabular} & \begin{tabular}[c]{@{}l@{}}Closed-mouth\\ Pistachios in\\ Testing Set\end{tabular} \\ \hline
Fold 1                                                & 339                                                        & 84                                                          & 1600                                                                              & 1550                                                                                & 393                                                                                 & 384                                                                                   \\ \hline
Fold 2                                                & 339                                                        & 84                                                          & 1610                                                                              & 1572                                                                                & 383                                                                                 & 362                                                                                   \\ \hline
Fold 3                                                & 339                                                        & 84                                                          & 1553                                                                              & 1506                                                                                & 440                                                                                 & 428                                                                                   \\ \hline
Fold 4                                                & 339                                                        & 84                                                          & 1641                                                                              & 1575                                                                                & 352                                                                                 & 359                                                                                   \\ \hline
Fold 5                                                & 336                                                        & 87                                                          & 1568                                                                              & 1533                                                                                & 425                                                                                 & 401                                                                                   \\ \hline
\end{tabular}

\label{table3}
\end{table*}

\subsection{Counting}
\label{33}
The second and main phase of our work was counting the number of open-mouth and closed-mouth pistachios in the videos. 
There were several challenges in this phase. The first challenge was that we wanted to develop a counting method that could be performed very fast on the CPU. Some of the other ideas may need a GPU; otherwise, the process would become highly time-consuming. However, our method works very much fast on the CPU, even faster than many other methods on GPU. As the object detection part needs a GPU for running in real-time mode, by applying a counting model on the CPU, the process of detection and counting can be executed simultaneously and much faster with no need to providing more expensive hardware. 

The second challenge was that some of the open-mouth pistachios could show themselves as closed-mouth pistachio in some consecutive frames and then reveal their open part only a few frames.  Moreover, some open-mouth pistachios rolling on the transportation line could show their open part several times and then appear like closed-mouth pistachios like Fig. \ref{fig6}.  We had to develop our algorithm somehow to prevent failing because of these challenges. 

Another challenge is developing the counting method to prevent failure because of false detections or not-detected pistachios that may be affected by the pistachios' occlusion.

For counting the pistachios, we first generate the frames from the taken video and then use the trained network for getting the bounding boxes of the pistachios in the frames. After this process, we would have a list of bounding boxes from the video's frames. 

Two thresholds have been designed to improve the counting accuracy: the initial threshold and the end threshold. The initial threshold is set to detect the newly inserted pistachios, and the end threshold is to reject adding the pistachios to the tracks list. These two methods are explained in the next paragraphs.

The algorithm first runs a function to set the initial threshold. In this function, the algorithm begins to assign the pistachios between each of two consecutive frames based on their distance without considering the class of pistachios. The minimum acceptable distance to assign the pistachios has been set to 20 pixels. The number of 20 pixels is taken out of our experience to prevent false assignments.
After the assigning,  the function adds the not-assigned pistachios that the height of the mid-point of their bounding box be less than 200 to a list (the height of the images is 600 pixels). These added pistachios are candidates as new inputs. After adding all the eligible pistachios, the function measures the average of the list, and it would be set as the initial threshold. This process performs to measure the area that most of the pistachios will enter the frame. The pistachios in different videos can enter the video frames differently and also may have various speeds, so this function will set the initial threshold separately for each video to improve the counting performance.

As the height of our image is 600 pixels, we set the end threshold equal to 500. We call the area between the top of an image (y=0) to the initial threshold (indicated by the counting model), the entering area, and the area between the end threshold (y=500) to the end of the image (y=600) the exiting area. The entering area is the main region for adding new incoming pistachios, but the exiting area includes pistachios that have been counted before, so we ignore them.

In the next level, the algorithm uses the assigned pistachios of each of the two consecutive frames that were computed in the last step.
Our counter algorithm role is to count the number of all pistachios and count the number of open-mouth and closed-mouth pistachios. The algorithm uses the initial threshold and the end threshold, which are computed in the last step. Toward solving the main challenge, which is that many of the pistachios may show their open part in some frames and the closed parts in the other frames, we have to track them by assigning them from frame to frame. We decided to track them from when they enter the entering area until they enter the exiting area. By doing this, we can know if one pistachio is open-mouth or closed-mouth, and it also prevents the algorithm from counting extra open-mouth pistachios.

Our algorithm analyzes each of the assigned pistachios in the two consecutive frames for all the frames. 
If the pistachio in the last frame was in the existing area or this pistachio in the current frame is in the exiting area,  this pistachio would also be rejected to be added to the track list. Otherwise, the pistachio in the current frame would be added to the track list that its assigned pistachio in the previous frame exists in it.

 After adding all the assigned pistachio in the current frame to the track lists that they belong to, the algorithm investigates the pistachios in the current frame that have not been assigned to any other pistachios. If these pistachios are located in the exiting area, they would be rejected for adding to any track list. If they be in the entering area and the number of all the pistachios in the current frame be greater than the number of pistachios in the last frame, these pistachios would be considered as new inputs and would start their own track list. The reason the pistachios in the current frame must be higher than the pistachios in the last frame is that in most cases, if a new pistachio enters the frame, the number of all pistachios should increase, but you may think that this situation may not always happen. It is true, but it also equalizes the conditions that some pistachios in the entering area will not be considered new inputs several times mistakenly.
 The unassigned pistachios in the current frame that can not be chosen as new inputs would be added to the Lost-Pistachios list. 
 
\begin{figure}[!hp]
\centering
\includegraphics[scale=0.48]{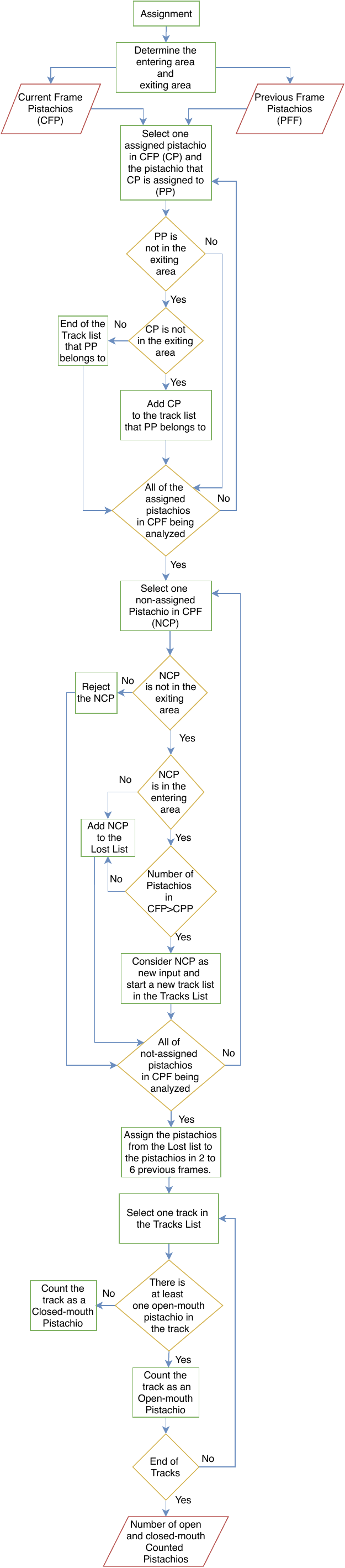}
\caption{The flow chart of the proposed counting algorithm}
\label{fig7}
\end{figure}

The Lost-Pistachios list is created to assign the pistachios that could not be assigned to the last frame pistachios (maybe because the pistachios in the last frame are not detected) to the pistachios in the 2 to 6 previous frames. 
 If the assignment is successful, the newly assigned pistachio will be added to the track list, and if not, they will be rejected.

Finally, after repeating this procedure for all the consecutive frames, we would have a list of tracked pistachios. If there be an open-mouth pistachio in a track, the whole track will be considered open-mouth. Therefore we could count the open-mouth and closed-mouth pistachios. The flow chart of the proposed counting algorithm is presented in Fig. \ref{fig7}.

\begin{figure}[!ht]
\centering
\includegraphics[scale=0.6]{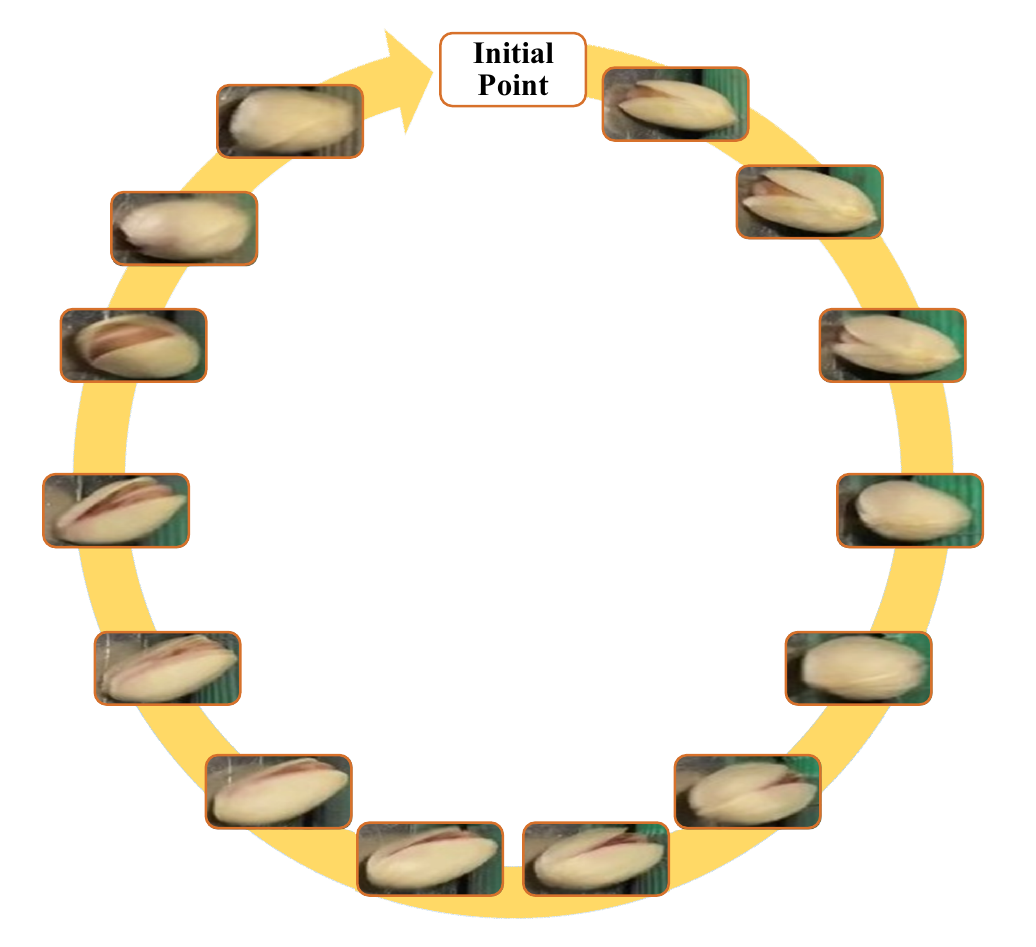}
\caption{In this figure, you can observe that a pistachio can be presented as open-mouth and closed-mouth several times while moving.}
\label{fig6}
\end{figure}

\newpage

\section{Results}
\label{4}
\subsection{Detection Results}
\label{41}
We trained Retinanet with three different backbones: ResNet50, ResNet152, and VGG16 based on the explained parameters in Table. \ref{table3} for 50 epochs. Data augmentation methods like rotation, translation, shearing, horizontal and vertical flipping, and rescaling were also applied to improve the training and prevent overfitting.

The system we used in this paper was provided by \href{https://colab.research.google.com/}{Google Colaboratory Notebooks}, which allocated a Tesla P100 GPU, 2.00GHz Intel Xeon CPU, and 25GB RAM on Linux to us. For utilizing RetinaNet we used the written codes by \href{https://github.com/fizyr}{Fizyr} which implemented RetinaNet with Keras library \cite{chollet2015keras} on Tensorflow backend \cite{tensorflow2015-whitepaper}.
The metrics we used for evaluating RetinaNet in the detection phase are Recall, Precision, F1 score, Accuracy, Average Precision(AP), and Mean Average Precision(mAP). AP is defined as:
\begin{equation}
AP = \frac{\sum_{i=1}^{D}\left \{ Precision(i)\times Recall(i) \right \}}{Number\ of\ annotations}\label{eq:5} 
\end{equation}

\begin{figure}[!ht]
\centering
\includegraphics[width=0.49\linewidth]{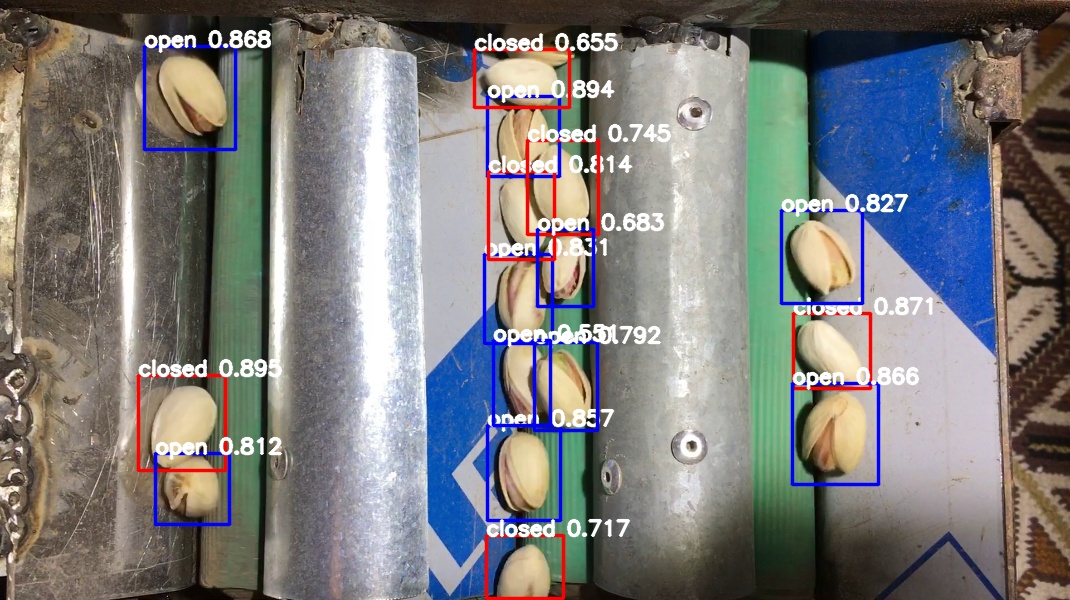}
\includegraphics[width=0.49\linewidth]{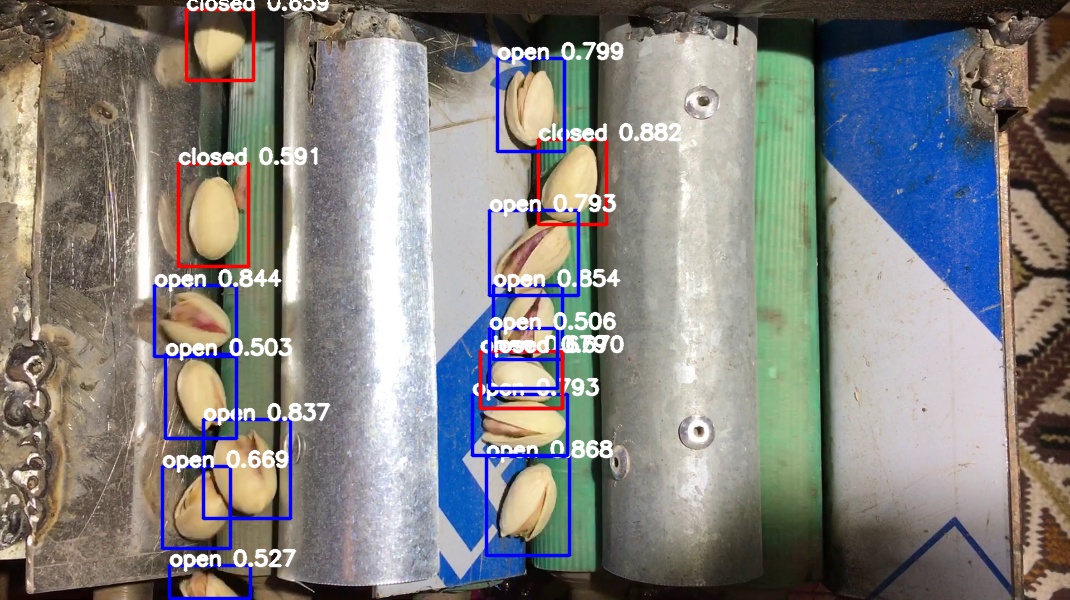}

\includegraphics[width=0.49\linewidth]{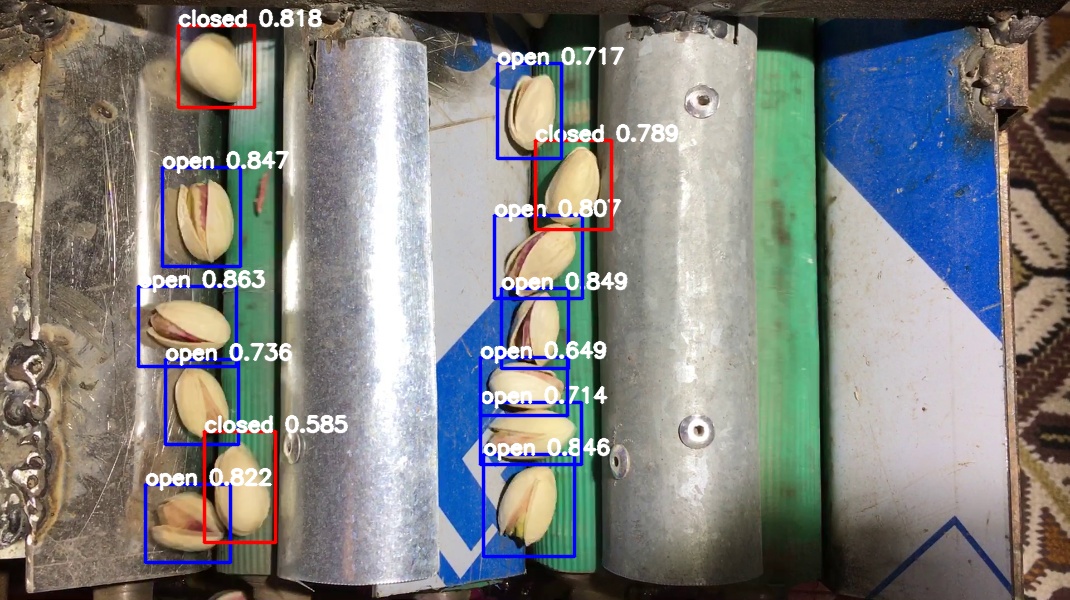}
\includegraphics[width=0.49\linewidth]{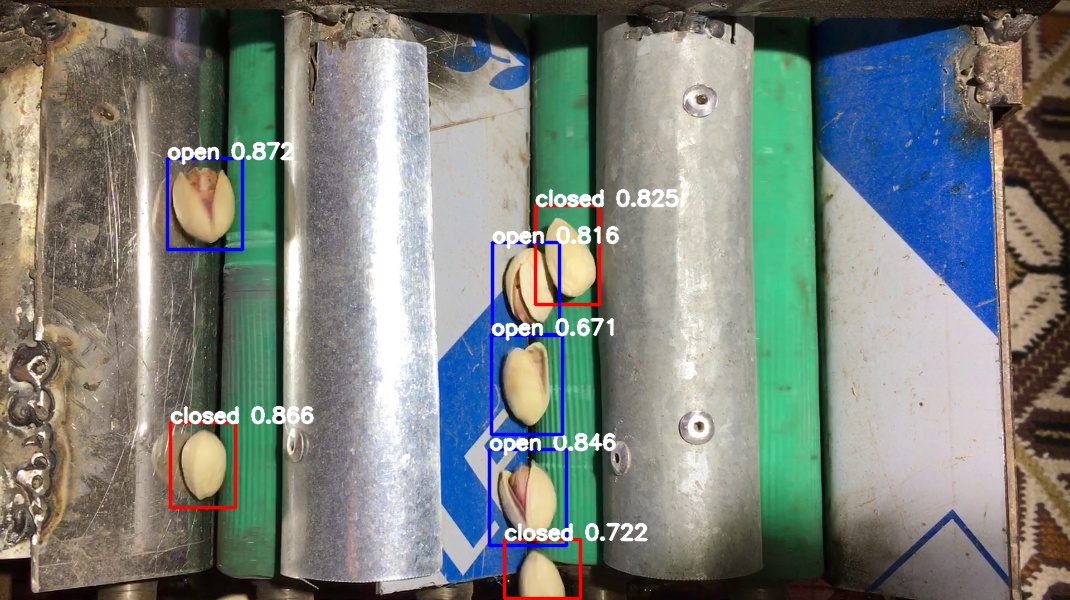}
\caption{These images are the output of RetinaNet. The red boxes are the closed-mouth pistachios, and the blue boxes belong to the open-mouth pistachios. The number beside the open or closed is the detection score value.}
\label{fig10}
\end{figure}

In Eq. \ref{eq:5}, D is the number of detected pistachios that are sorted by scores. Average Precision will be calculated for each class separately.
The mAP metric is the mean of Average Precision between classes.
Fig. \ref{fig10} presents some of the images with the detected pistachios.

We considered the detected boxes with Intersection over Union (IOU) more than 0.5 as true positives and the others as false positives to evaluate the detection results.
The detection results of RetinaNet on the three backbones are presented in Tables. \ref{detection_claases} and \ref{detection-all}.

Table. \ref{detection-all}. shows that the performance of ResNet backbones is better than VGG16, as expected. The average mAP between five-folds for VGG16, ResNet152, and ResNet50 is 91.23\%; 91.69\%, and 91.87\% respectively. The performance of ResNet50 and ResNet152 is nearly close. Understandably, as there are only two classes in this dataset, and pistachios attributes are distinguishable from the background,  there is no need to use bigger models. Using bigger models in our dataset may lead to overfitting and does not increase the output mAP. The VGG16 model is much weaker in image classification tasks than the ResNet models on the ImageNet dataset. Still, RetinaNet on the VGG16 backbone model obtained good detection results, which shows the high effect of feature pyramid network and focal loss.

\begin{table*}[!hb]
\centering
\large
\caption{In this table, the processing time of our counter algorithm after getting the detections from RetinaNet is reported for each video.}
\begin{tabular}{|l|l|l|}
\hline
Video  & \begin{tabular}[c]{@{}l@{}}Number\\ of\\ Frames\end{tabular} & Time (S)                    \\ \hline
Video1 & {\color[HTML]{000000} 984}                                   & {\color[HTML]{000000} 3.00} \\ \hline
Video2 & {\color[HTML]{000000} 1665}                                  & {\color[HTML]{000000} 3.39} \\ \hline
Video3 & {\color[HTML]{000000} 1833}                                  & {\color[HTML]{000000} 5.11} \\ \hline
Video4 & {\color[HTML]{000000} 2227}                                  & {\color[HTML]{000000} 5.75} \\ \hline
Video5 & {\color[HTML]{000000} 2171}                                  & {\color[HTML]{000000} 4.19} \\ \hline
Video6 & {\color[HTML]{000000} 606}                                   & {\color[HTML]{000000} 0.83} \\ \hline
\end{tabular}

\label{speed}
\end{table*}

\begin{table*}[!ht]
\centering
\large
\caption{The detection results of RetinaNet on different backbones for each fold are reported in this table.}
\begin{adjustbox}{width=1\textwidth}
\begin{tabular}{ll|lll|lll|}
\cline{3-8}
                            &           &                                                    & \begin{tabular}[c]{@{}c@{}}Class\\ closed-pistachio\end{tabular} &                               &                                                    & \begin{tabular}[c]{@{}c@{}}Class\\ open-pistachio\end{tabular} &                               \\ \cline{3-8}
\multicolumn{1}{c}{}      &           & \multicolumn{1}{c|}{AP}                            & \multicolumn{1}{c|}{F1 score}                                    & Recall                        & \multicolumn{1}{c|}{AP}                            & \multicolumn{1}{c|}{F1 score}                                  & Recall                        \\ \cline{1-8} 
\multicolumn{1}{|l|}{}      & ResNet50  & \multicolumn{1}{l|}{{\color[HTML]{000000} 0.9072}} & \multicolumn{1}{l|}{{\color[HTML]{000000} 0.9128}}               & {\color[HTML]{000000} 0.9270} & \multicolumn{1}{l|}{{\color[HTML]{000000} 0.9467}} & \multicolumn{1}{l|}{{\color[HTML]{000000} 0.9325}}             & {\color[HTML]{000000} 0.9669} \\ \cline{2-8} 
\multicolumn{1}{|l|}{Fold1} & ResNet152 & \multicolumn{1}{l|}{{\color[HTML]{000000} 0.9037}} & \multicolumn{1}{l|}{{\color[HTML]{000000} 0.9238}}               & {\color[HTML]{000000} 0.9166} & \multicolumn{1}{l|}{{\color[HTML]{000000} 0.9686}} & \multicolumn{1}{l|}{{\color[HTML]{000000} 0.9302}}             & {\color[HTML]{000000} 0.9847} \\ \cline{2-8} 
\multicolumn{1}{|l|}{}      & VGG16     & \multicolumn{1}{l|}{{\color[HTML]{000000} 0.8821}} & \multicolumn{1}{l|}{{\color[HTML]{000000} 0.9067}}               & {\color[HTML]{000000} 0.8984} & \multicolumn{1}{l|}{{\color[HTML]{000000} 0.976}}  & \multicolumn{1}{l|}{{\color[HTML]{000000} 0.9397}}             & {\color[HTML]{000000} 0.9923} \\ \hline
\multicolumn{1}{|l|}{}      & ResNet50  & \multicolumn{1}{l|}{{\color[HTML]{000000} 0.9135}} & \multicolumn{1}{l|}{{\color[HTML]{000000} 0.9213}}               & {\color[HTML]{000000} 0.9226} & \multicolumn{1}{l|}{{\color[HTML]{000000} 0.9172}} & \multicolumn{1}{l|}{{\color[HTML]{000000} 0.9359}}             & {\color[HTML]{000000} 0.9347} \\ \cline{2-8} 
\multicolumn{1}{|l|}{Fold2} & ResNet152 & \multicolumn{1}{l|}{{\color[HTML]{000000} 0.9010}} & \multicolumn{1}{l|}{{\color[HTML]{000000} 0.9105}}               & {\color[HTML]{000000} 0.9143} & \multicolumn{1}{l|}{{\color[HTML]{000000} 0.9151}} & \multicolumn{1}{l|}{{\color[HTML]{000000} 0.9286}}             & {\color[HTML]{000000} 0.9347} \\ \cline{2-8} 
\multicolumn{1}{|l|}{}      & VGG16     & \multicolumn{1}{l|}{{\color[HTML]{000000} 0.8841}} & \multicolumn{1}{l|}{{\color[HTML]{000000} 0.8856}}               & {\color[HTML]{000000} 0.9198} & \multicolumn{1}{l|}{{\color[HTML]{000000} 0.9157}} & \multicolumn{1}{l|}{{\color[HTML]{000000} 0.8988}}             & {\color[HTML]{000000} 0.9399} \\ \hline
\multicolumn{1}{|l|}{}      & ResNet50  & \multicolumn{1}{l|}{{\color[HTML]{000000} 0.9402}} & \multicolumn{1}{l|}{{\color[HTML]{000000} 0.9232}}               & {\color[HTML]{000000} 0.9556} & \multicolumn{1}{l|}{{\color[HTML]{000000} 0.8832}} & \multicolumn{1}{l|}{{\color[HTML]{000000} 0.9130}}             & {\color[HTML]{000000} 0.9068} \\ \cline{2-8} 
\multicolumn{1}{|l|}{Fold3} & ResNet152 & \multicolumn{1}{l|}{{\color[HTML]{000000} 0.9096}} & \multicolumn{1}{l|}{{\color[HTML]{000000} 0.9161}}               & {\color[HTML]{000000} 0.9322} & \multicolumn{1}{l|}{{\color[HTML]{000000} 0.8981}} & \multicolumn{1}{l|}{{\color[HTML]{000000} 0.9171}}             & {\color[HTML]{000000} 0.9181} \\ \cline{2-8} 
\multicolumn{1}{|l|}{}      & VGG16     & \multicolumn{1}{l|}{{\color[HTML]{000000} 0.8919}} & \multicolumn{1}{l|}{{\color[HTML]{000000} 0.9138}}               & {\color[HTML]{000000} 0.9042} & \multicolumn{1}{l|}{{\color[HTML]{000000} 0.8922}} & \multicolumn{1}{l|}{{\color[HTML]{000000} 0.9032}}             & {\color[HTML]{000000} 0.9227} \\ \hline
\multicolumn{1}{|l|}{}      & ResNet50  & \multicolumn{1}{l|}{{\color[HTML]{000000} 0.9183}} & \multicolumn{1}{l|}{{\color[HTML]{000000} 0.9115}}               & {\color[HTML]{000000} 0.9331} & \multicolumn{1}{l|}{{\color[HTML]{000000} 0.9301}} & \multicolumn{1}{l|}{{\color[HTML]{000000} 0.9258}}             & {\color[HTML]{000000} 0.9403} \\ \cline{2-8} 
\multicolumn{1}{|l|}{Fold4} & ResNet152 & \multicolumn{1}{l|}{{\color[HTML]{000000} 0.9347}} & \multicolumn{1}{l|}{{\color[HTML]{000000} 0.9243}}               & {\color[HTML]{000000} 0.9526} & \multicolumn{1}{l|}{{\color[HTML]{000000} 0.9449}} & \multicolumn{1}{l|}{{\color[HTML]{000000} 0.9318}}             & {\color[HTML]{000000} 0.9517} \\ \cline{2-8} 
\multicolumn{1}{|l|}{}      & VGG16     & \multicolumn{1}{l|}{{\color[HTML]{000000} 0.9186}} & \multicolumn{1}{l|}{{\color[HTML]{000000} 0.9110}}               & {\color[HTML]{000000} 0.9415} & \multicolumn{1}{l|}{{\color[HTML]{000000} 0.9303}} & \multicolumn{1}{l|}{{\color[HTML]{000000} 0.9194}}             & {\color[HTML]{000000} 0.9403} \\ \hline
\multicolumn{1}{|l|}{}      & ResNet50  & \multicolumn{1}{l|}{{\color[HTML]{000000} 0.9122}} & \multicolumn{1}{l|}{{\color[HTML]{000000} 0.9106}}               & {\color[HTML]{000000} 0.9276} & \multicolumn{1}{l|}{{\color[HTML]{000000} 0.9179}} & \multicolumn{1}{l|}{{\color[HTML]{000000} 0.9175}}             & {\color[HTML]{000000} 0.9294} \\ \cline{2-8} 
\multicolumn{1}{|l|}{Fold5} & ResNet152 & \multicolumn{1}{l|}{{\color[HTML]{000000} 0.8830}} & \multicolumn{1}{l|}{{\color[HTML]{000000} 0.8996}}               & {\color[HTML]{000000} 0.9052} & \multicolumn{1}{l|}{{\color[HTML]{000000} 0.9100}} & \multicolumn{1}{l|}{{\color[HTML]{000000} 0.9160}}             & {\color[HTML]{000000} 0.9247} \\ \cline{2-8} 
\multicolumn{1}{|l|}{}      & VGG16     & \multicolumn{1}{l|}{{\color[HTML]{000000} 0.8922}} & \multicolumn{1}{l|}{{\color[HTML]{000000} 0.8907}}               & {\color[HTML]{000000} 0.9152} & \multicolumn{1}{l|}{{\color[HTML]{000000} 0.8979}} & \multicolumn{1}{l|}{{\color[HTML]{000000} 0.9153}}             & {\color[HTML]{000000} 0.9035} \\ \hline
\end{tabular}
\end{adjustbox}
\label{detection_claases}
\end{table*}

\begin{table*}[!ht]
\centering
\large
\caption{This table contains the RetinaNet evaluation data for all the classes.}
\begin{adjustbox}{width=1\textwidth}
\begin{tabular}{l|l|l|l|l|l|l|l|l|l|}
\cline{2-10}
                              & Backbone Network   & TP                         & FP                         & FN                        & Recall                                 & Precision                              & F1 score                               & mAP                                     & Accuracy                               \\ \hline
\multicolumn{1}{|l|}{}        & ResNet50  & {\color[HTML]{000000} 736} & {\color[HTML]{000000} 82}  & {\color[HTML]{000000} 41} & {\color[HTML]{000000} 0.9472}          & {\color[HTML]{000000} 0.8997}          & {\color[HTML]{000000} 0.9228}          & {\color[HTML]{000000} 0.9270}         & {\color[HTML]{000000} 0.8568}         \\ \cline{2-10} 
\multicolumn{1}{|l|}{Fold1}   & ResNet152 & {\color[HTML]{000000} 739} & {\color[HTML]{000000} 78}  & {\color[HTML]{000000} 38} & {\color[HTML]{000000} \textbf{0.9510}} & {\color[HTML]{000000} \textbf{0.9045}} & {\color[HTML]{000000} \textbf{0.9272}} & {\color[HTML]{000000} \textbf{0.9361}} & {\color[HTML]{000000} \textbf{0.8643}} \\ \cline{2-10} 
\multicolumn{1}{|l|}{}        & VGG16     & {\color[HTML]{000000} 735} & {\color[HTML]{000000} 79}  & {\color[HTML]{000000} 42} & {\color[HTML]{000000} 0.9459}          & {\color[HTML]{000000} 0.9029}          & {\color[HTML]{000000} 0.9239}          & {\color[HTML]{000000} 0.9291}           & {\color[HTML]{000000} 0.8586}          \\ \hline
\multicolumn{1}{|l|}{}        & ResNet50  & {\color[HTML]{000000} 692} & {\color[HTML]{000000} 53}  & {\color[HTML]{000000} 53} & {\color[HTML]{000000} 0.9288}          & {\color[HTML]{000000} \textbf{0.9288}} & {\color[HTML]{000000} \textbf{0.9288}} & {\color[HTML]{000000} \textbf{0.9153}}  & {\color[HTML]{000000} \textbf{0.8671}} \\ \cline{2-10} 
\multicolumn{1}{|l|}{Fold2}   & ResNet152 & {\color[HTML]{000000} 689} & {\color[HTML]{000000} 64}  & {\color[HTML]{000000} 56} & {\color[HTML]{000000} 0.9248}          & {\color[HTML]{000000} 0.9150}          & {\color[HTML]{000000} 0.9198}          & {\color[HTML]{000000} 0.90812}          & {\color[HTML]{000000} 0.8516}          \\ \cline{2-10} 
\multicolumn{1}{|l|}{}        & VGG16     & {\color[HTML]{000000} 693} & {\color[HTML]{000000} 115} & {\color[HTML]{000000} 52} & {\color[HTML]{000000} \textbf{0.9302}} & {\color[HTML]{000000} 0.8576}          & {\color[HTML]{000000} 0.8924}          & {\color[HTML]{000000} 0.8999}           & {\color[HTML]{000000} 0.8058}          \\ \hline
\multicolumn{1}{|l|}{}        & ResNet50  & {\color[HTML]{000000} 808} & {\color[HTML]{000000} 84}  & {\color[HTML]{000000} 60} & {\color[HTML]{000000} \textbf{0.9308}} & {\color[HTML]{000000} 0.9058}          & {\color[HTML]{000000} \textbf{0.9181}} & {\color[HTML]{000000} \textbf{0.9117}}  & {\color[HTML]{000000} \textbf{0.8487}} \\ \cline{2-10} 
\multicolumn{1}{|l|}{Fold3}   & ResNet152 & {\color[HTML]{000000} 803} & {\color[HTML]{000000} 81}  & {\color[HTML]{000000} 65} & {\color[HTML]{000000} 0.9251}          & {\color[HTML]{000000} \textbf{0.9083}} & {\color[HTML]{000000} 0.9166}          & {\color[HTML]{000000} 0.9038}           & {\color[HTML]{000000} 0.8461}          \\ \cline{2-10} 
\multicolumn{1}{|l|}{}        & VGG16     & {\color[HTML]{000000} 793} & {\color[HTML]{000000} 85}  & {\color[HTML]{000000} 75} & {\color[HTML]{000000} 0.9135}          & {\color[HTML]{000000} 0.9031}          & {\color[HTML]{000000} 0.9083}          & {\color[HTML]{000000} 0.8921}           & {\color[HTML]{000000} 0.8321}          \\ \hline
\multicolumn{1}{|l|}{}        & ResNet50  & {\color[HTML]{000000} 666} & {\color[HTML]{000000} 73}  & {\color[HTML]{000000} 45} & {\color[HTML]{000000} 0.9367}          & {\color[HTML]{000000} 0.9012}          & {\color[HTML]{000000} 0.9186}          & {\color[HTML]{000000} 0.9242}           & {\color[HTML]{000000} 0.8494}          \\ \cline{2-10} 
\multicolumn{1}{|l|}{Fold4}   & ResNet152 & {\color[HTML]{000000} 677} & {\color[HTML]{000000} 71}  & {\color[HTML]{000000} 34} & {\color[HTML]{000000} \textbf{0.9521}} & {\color[HTML]{000000} \textbf{0.9050}} & {\color[HTML]{000000} \textbf{0.9280}} & {\color[HTML]{000000} \textbf{0.9398}}  & {\color[HTML]{000000} \textbf{0.8657}} \\ \cline{2-10} 
\multicolumn{1}{|l|}{}        & VGG16     & {\color[HTML]{000000} 669} & {\color[HTML]{000000} 82}  & {\color[HTML]{000000} 42} & {\color[HTML]{000000} 0.9409}          & {\color[HTML]{000000} 0.8908}          & {\color[HTML]{000000} 0.9151}          & {\color[HTML]{000000} 0.9244}           & {\color[HTML]{000000} 0.8436}          \\ \hline
\multicolumn{1}{|l|}{}        & ResNet50  & {\color[HTML]{000000} 767} & {\color[HTML]{000000} 85}  & {\color[HTML]{000000} 59} & {\color[HTML]{000000} \textbf{0.9285}} & {\color[HTML]{000000} 0.9002}          & {\color[HTML]{000000} \textbf{0.9141}} & {\color[HTML]{000000} \textbf{0.9151}}  & {\color[HTML]{000000} \textbf{0.8419}} \\ \cline{2-10} 
\multicolumn{1}{|l|}{Fold5}   & ResNet152 & {\color[HTML]{000000} 756} & {\color[HTML]{000000} 83}  & {\color[HTML]{000000} 70} & {\color[HTML]{000000} 0.9152}          & {\color[HTML]{000000} \textbf{0.9010}} & {\color[HTML]{000000} 0.9081}          & {\color[HTML]{000000} 0.8965}           & {\color[HTML]{000000} 0.8316}          \\ \cline{2-10} 
\multicolumn{1}{|l|}{}        & VGG16     & {\color[HTML]{000000} 751} & {\color[HTML]{000000} 86}  & {\color[HTML]{000000} 75} & {\color[HTML]{000000} 0.9092}          & {\color[HTML]{000000} 0.8972}          & {\color[HTML]{000000} 0.9031}          & {\color[HTML]{000000} 0.8950}           & {\color[HTML]{000000} 0.8234}          \\ \hline
\multicolumn{1}{|l|}{}        & ResNet50  & \textbf{733.8}             & \textbf{75.4}              & \textbf{51.6}                      & \textbf{0.9344}                        & \textbf{0.9071}                        & \textbf{0.9205}                        & \textbf{0.9187}                         & \textbf{0.8528}                        \\ \cline{2-10} 
\multicolumn{1}{|l|}{Average} & ResNet152 & 732.8                      & \textbf{75.4}              & 52.6                      & 0.9336                                 & 0.9068                                 & 0.9199                                 & 0.9169                                  & 0.8519                                 \\ \cline{2-10} 
\multicolumn{1}{|l|}{}        & VGG16     & 728.2                      & 89.4                       & 57.2             & 0.9279                                 & 0.8903                                 & 0.9086                                 & 0.9123                                  & 0.8332                                 \\ \hline
\end{tabular}
\end{adjustbox}

\label{detection-all}
\end{table*}

\subsection{Counting Results}
\label{412}

Six different videos with 167 seconds length and 9486 frames were selected for evaluating the counting algorithm. We tested our counting algorithm based on the detections gathered from the trained networks on different backbones.
The results and the overall accuracy for all the videos are present in Table. \ref{summary} and Table. \ref{speed} also expresses the running time of the counting model.

In Table. \ref{summary}, the accuracy metric is considered for evaluating the tracking algorithm, which is defined as

\begin{equation}
Accuracy=\frac{TP}{TP+FN+FP}\label{eq:1}
\end{equation}
In equation \ref{eq:1}, TP is the number of the correct-counted pistachios, FN is the number of not-counted pistachios, and FP is the number of extra miscounted pistachios. 

The information in Table. \ref{summary}, have been calculated based on the trained models in our dataset's first fold. As ResNet152 got better results in Fold1 (Table. \ref{detection-all}), we can see that the counting results are better for this model in Fold1. The counting accuracy based on the detections gathered from RetinaNet on ResNet152; ResNet50 and VGG16 backbones are 94.75\%, 91.96\%, and 90.96\%, respectively. The speed of the counting model (Table. \ref{speed}) is also reasonable compared to the number frames; e.g., video4 includes 2227 images, and the counting model only takes 5.75 seconds to run. It must be noted that this time is the processing time of just the counting model and does not include object detector processing latency.

\begin{table*}[!ht]
\centering
\large
\caption{Counting results for all the 6 test videos. The detections are taken from the trained networks in the first fold. Extra counted means the sum of miscounted open-mouth and closed-mouth pistachios.}
\begin{adjustbox}{width=1\textwidth}
\begin{tabular}{|l|l|l|l|l|l|l|}
\hline
Backbone Network   & \begin{tabular}[c]{@{}l@{}}Ground-Truth\\ Open-Mouth\\ Pistachios\end{tabular} & \begin{tabular}[c]{@{}l@{}}Ground-Truth\\ Closed-Mouth\\ Pistachios\end{tabular} & \begin{tabular}[c]{@{}l@{}}Correctly\\ Counted\\ Open-Mouth\\ Pistachios\end{tabular} & \begin{tabular}[c]{@{}l@{}}Correctly\\ Counted\\ Closed-Mouth\\ Pistachios\end{tabular} & \begin{tabular}[c]{@{}l@{}}Extra\\ Counted\end{tabular} & Accuracy        \\ \hline
ResNet152 & 409                                                                            & 152                                                                              & 397                                                                                   & 145                                                                                     & 11                                                      & \textbf{0.9475} \\ \hline
ResNet50  & 409                                                                            & 152                                                                              & 386                                                                                   & 152                                                                                     & 24                                                      & 0.9196          \\ \hline
VGG16     & 409                                                                            & 152                                                                              & 395                                                                                   & 149                                                                                     & 37                                                      & 0.9096          \\ \hline
\end{tabular}
\end{adjustbox}

\label{summary}
\end{table*}
\begin{figure*}[!ht]
\centering
\includegraphics[scale=0.55]{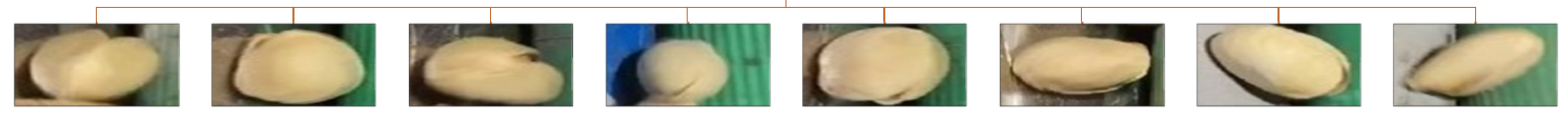}
\caption{Some of the examples in our dataset which are hard to classify}
\label{fig8}
\end{figure*}

\section{Discussion}
\label{5}
Based on the Table. \ref{detection-all}, in fold 1 and 4, ResNet152 performs better, but in other folds, ResNet50 achieves better results. On average between five-folds ResNet50 and ResNet152 get 91.87\% and 91.69\% mAP respectively.

It is clear that RetinaNet on ResNet models achieves better results, but the performance of ResNet50 and ResNet152 are almost equal. 
Although; ResNet152 is a deeper model, but in this dataset, as pistachios' features are apparent from the background and there are only two classes, there is no need to use bigger models, and the ResNet50 model is sufficient in this case. 

In Fig. \ref{fig8}, it can be visualized that the open-mouth and the closed-mouth pistachios could look like each other, and it would be hard to distinguish them even with human eyes. So based on the detection mAP and accuracy, the models are robust for detecting the pistachios.

Our counting model's accuracy on ResNet152 backbone in fold1 is 94.75\% while the detection accuracy on ResNet152 backbone in Fold1 is 86.43\%. Also from Table. \ref{summary}, we can see that the counting results are promising and are even higher than the detection results. This shows our proposed counter model's robustness and its ability to compensate the object detector's failures and that it can assign and track the pistachios within the video frames with high accuracy.
Our counting model has been evaluated on six different videos containing 9486 frames with 561 moving pistachios and more than 350,000 single pistachios (sum of pistachios in all frames). 
Based on Table. \ref{speed}, the counting procedure has been performed fast. This algorithm can be utilized in factories and industries related to pistachios because of its high speed and good accuracy.

\section{Conclusion}
\label{6}

Pistachio is a nutritious nut that originated from central Asia and the middle east and is mainly sorted into open-mouth and closed-mouth kinds. Open-mouth pistachios worth much more than closed-mouth pistachios, so it can be very valuable for the related companies to know about the exact amount of these two kinds in their packages.
This paper has proposed a novel model that can be used to design a remote AI system to detect and count the number of open-mouth and closed-mouth pistachios in the factories' production line. 
We have introduced a new dataset that is called Pesteh-Set. It is made of 6 videos (9486 frames) and 3927 labeled open-mouth and closed-mouth pistachios.  

Pistachios are motile objects and usually spin on the transportation line; therefore, from the view of a camera, open-mouth pistachios may place on their backside and appear like closed-mouth pistachios and again appear as open-mouth in some other frames. Due to this challenge, we had to develop our methods in a way to prevent false counting.

Our proposed model is constructed of a detection part and a counting part. For the detection part, we have implemented the RetinaNet object detector on our dataset to detect the different types of pistachios in video frames. We also trained and investigated RetinaNet on three different backbones: ResNet50, ResNet152, and VGG16 to find the best-performing model. The Mean Average Precision (mAP) between five-folds for RetinaNet on the ResNet50 network was 91.87\%. 

Our developed counter model; first receives the detected pistachios from the object detector model and then applies our designed tracking method to assign pistachios in consecutive frames. Our tracker's most crucial point is its ability to track open-mouth pistachios that appear as closed-mouth in some frames, even in pistachios' high density and occlusion.
Our counter model performs fast, with no need for GPU (besides the object detection part), and achieves good results. It was tested on six different videos containing 9486 frames with 561 moving pistachios and more than 350,000 single pistachios (sum of pistachios in all frames).
This algorithm obtained 94.75\% accuracy.

This work can be extended to detect and count more than two classes of pistachios, e.g., .semi-open mouth pistachios can also be added to the classes.
The detection accuracy can also be increased in future works. One way is that researchers can use our developed program and the dataset to generate more labeled images. They can label all the frames of some videos and use this method \cite{rahimzadeh2020sperm} that was proposed to improve the motile-objects detection. Pistachios are motile objects, so by using this method, the detection accuracy should be improved.

\section{Data Availability}
\label{7}
In this GitHub profile (\url{https://github.com/mr7495/Pesteh-Set}), we have shared our dataset and all the codes that were used for preparing and labeling the dataset.

\section{Code Availability}
\label{8}
In this GitHub profile (\url{https://github.com/mr7495/Pistachio-Counting}), we made the trained neural networks, the counting algorithm and all the codes that were used for training and validating the networks, public for researchers use.

\section*{Acknowledgment}
We wish to thank Mr.Navid Akhundi, who recorded our dataset videos and shared them with us, and \href{https://github.com/fizyr}{Fizyr}, who implemented RetinaNet with Keras on GitHub. 
We also thank \href{https://colab.research.google.com/}{ Colab server} for providing free and powerful GPU and \href{https://drive.google.com/}{Google Drive} for providing space for data hosting.

This is a preprint of an article published in the Iran Journal of Computer Science. The final authenticated version is available online at \href{https://doi.org/10.1007/s42044-021-00090-6}{doi.org/10.1007/s42044-021-00090-6}

\section*{Conflict of Interests}

The authors declare that they have no known competing financial interests or personal relationships that could have appeared to influence the work reported in this paper.

\bibliographystyle{abbrv}
\bibliography{arxiv}

\end{document}